\documentclass[10pt,twocolumn,letterpaper]{article}

\usepackage[pagenumbers]{cvpr}      
\usepackage{amsmath}
\usepackage{multirow}
\usepackage{pifont}
\usepackage{makecell}
\usepackage{subcaption}

\usepackage{amsmath}
\usepackage{multirow}
\usepackage{pifont}
\usepackage{makecell}
\usepackage{subcaption}
\usepackage{colortbl} 

\setlist{nolistsep}

\newcommand{\methodname}[1]{\texttt{U-SAM}}

\definecolor{cvprblue}{rgb}{0.21,0.49,0.74}
\usepackage[pagebackref,breaklinks,colorlinks,citecolor=cvprblue]{hyperref}

\def\paperID{5727} 
\def\confName{CVPR}
\def\confYear{2024}

\title{Repurposing SAM for User-Defined Semantics Aware Segmentation\thanks{This paper has been published at The IEEE/CVF Conference on Computer Vision and Pattern Recognition 2025}}
\author{Rohit Kundu$^{1}$, Sudipta Paul$^{2}$, Arindam Dutta$^{1}$, Amit K. Roy-Chowdhury$^{1}$\\
$^{1}$University of California, Riverside ~$^{2}$Samsung Research America\\
{\tt \small \{rkund006@, spaul007@, adutt020@, amitrc@ece.\}ucr.edu}}

\begin{document}
\maketitle

\begin{abstract}
    The Segment Anything Model (SAM) excels at generating precise object masks from input prompts but lacks semantic awareness, failing to associate its generated masks with specific object categories.  To address this limitation, we propose \methodname{}, a novel framework that imbibes semantic awareness into SAM, enabling it to generate targeted masks for user-specified object categories. Given only object class names as input from the user, \methodname{} provides pixel-level semantic annotations for images without requiring any labeled/unlabeled samples from the test data distribution. Our approach leverages synthetically generated or web crawled images to accumulate semantic information about the desired object classes. We then learn a mapping function between SAM's mask embeddings and object class labels, effectively enhancing SAM with granularity-specific semantic recognition capabilities. As a result, users can obtain meaningful and targeted segmentation masks for specific objects they request, rather than generic and unlabeled masks. We evaluate \methodname{} on PASCAL VOC 2012 and MSCOCO-80, achieving significant $mIoU$ improvements of $+17.95\%$ and $+5.20\%$, respectively, over state-of-the-art methods. By transforming SAM into a semantically aware segmentation model, \methodname{} offers a practical and flexible solution for pixel-level annotation across diverse and unseen domains in a resource-constrained environment.
\end{abstract}

\section{Introduction}\label{sec:intro}

\begin{figure}[t]
    \centering
    \includegraphics[width=\columnwidth]{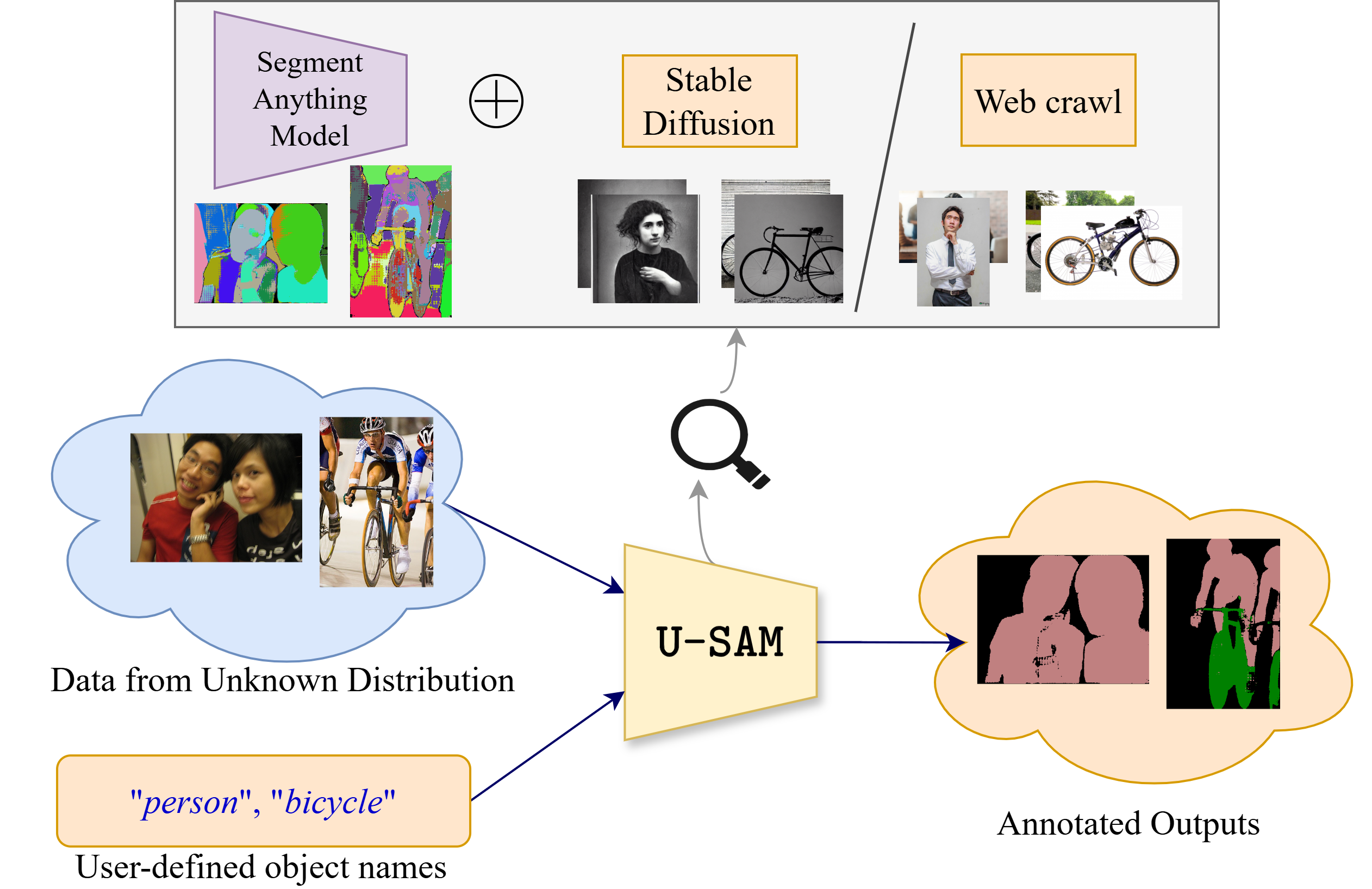}
    \caption{\textbf{Problem Overview}: We propose a novel and flexible pipeline, \methodname{}, to tackle the challenging problem of generating pixel-level semantic annotations for user-specified object classes without any manual supervision. Given only a set of object class names, \methodname{} can produce accurate semantic segmentation masks on any user-provided image data. By enhancing SAM with semantic region recognition capabilities and leveraging synthetic images generated by Stable Diffusion or web crawled images for the desired object classes, our approach addresses a highly complex segmentation problem with robust generalization.}
    \label{fig:problem_statement}
\end{figure}

\noindent
Semantic segmentation \cite{he2020momentum, basak2022mfsnet, caron2020unsupervised, basak2023ideal} has become a cornerstone of modern deep learning, enabling pixel-level understanding of images for applications in autonomous driving \cite{chen2023edge, yao2023radar}, satellite imaging \cite{yan2023semantic, li2023semanticremote}, and robotics \cite{yan2020roboseg, gong2023integrated}. However, its success heavily depends on large amounts of labeled data for supervised training \cite{you2023ferret, zhang2023text2seg}, which makes manual annotation both expensive and time-intensive. Labeling every pixel in an image to generate semantic masks is particularly laborious, creating a significant bottleneck.

The Segment Anything Model (SAM) \cite{kirillov2023segment} has emerged as a powerful foundation model for segmentation tasks, capable of generating precise object masks from input prompts. However, SAM's lack of semantic awareness restricts its utility in real-world applications where precise object identification is crucial. For instance, SAM can segment objects based on prompts like points or bounding boxes but cannot inherently distinguish between different object categories. This limitation means that SAM produces numerous unlabeled masks, requiring additional processing to identify specific objects.

To bridge this gap, we propose a novel approach that enhances SAM with the ability to generate semantic segmentation masks for user-specified object categories. Our method, \textit{\underline{U}ser-Defined Semantics Aware \underline{SAM}} or \methodname{}, requires only the names of the target object categories as input and outputs semantic segmentation masks for user-provided datasets during inference. Unlike existing methods that rely on labeled \cite{strudel2021segmenter, guo2022segnext} or unlabeled \cite{guo2021metacorrection, lee2021unsupervised, zhao2024unsupervised} data from the test dataset distribution, our approach eliminates the need for in-domain training data or prior knowledge of the test dataset distribution. This generalization allows users to provide images from any dataset, and our model reliably generates semantic masks tailored to the specified objects, such as ``person" or ``bicycle," as shown in Figure \ref{fig:problem_statement}.

Notably, most weakly supervised segmentation methods \cite{ru2023token, he2023weakly, yang2024foundation}, along with in-domain training data, also require object labels for every image. In contrast, our pipeline relaxes even this assumption by requiring only dataset-level object names. During inference, we do not know which images will contain which objects, making our problem setup more challenging than traditional unsupervised or weakly supervised learning and closer to zero-shot learning. This distinction underscores the novelty and complexity of our approach, which addresses a significant gap in current segmentation methodologies.

While generative models like Stable Diffusion \cite{rombach2022high} are capable of producing high-quality synthetic data, state-of-the-art semantic segmentation methods have yet to fully leverage such models for downstream tasks. So, to address this new segmentation challenge, we leverage synthetic images generated by Stable Diffusion \cite{rombach2022high} or web-crawled images corresponding to user-specified object categories. We then train a classifier head on these images to enhance SAM with semantic recognition capabilities. Once trained, \methodname{} can perform semantic segmentation on any dataset—whether it is PASCAL VOC \cite{pascalvoc2012}, MSCOCO-80 \cite{lin2014microsoft}, or a custom user-created dataset—without requiring retraining or fine-tuning for each new test distribution. This capability makes \methodname{} highly practical and versatile for real-world applications.

By transforming SAM \cite{kirillov2023segment} into a semantically aware model, we address a critical need in real-world applications where users require precise object identification without the burden of manual annotation or dataset-specific training. Our approach not only enhances SAM's capabilities but also offers a practical solution for pixel-level annotation across diverse and unseen domains.

\noindent In summary, our \textbf{contributions} are as follows:
\begin{itemize}
    \item We propose \methodname{}, a novel framework that enhances the SAM \cite{kirillov2023segment} by imbuing it with semantic awareness, allowing it to generate targeted semantic masks for user-specified object categories without manual supervision.

    \item Our approach eliminates the need for in-domain labeled/unlabeled training data from the test distribution, allowing \methodname{} to operate without prior dataset knowledge. We leverage synthetic images generated by Stable Diffusion \cite{rombach2022high} or web crawled images based on user-provided object class names, creating a flexible pipeline that generalizes across diverse datasets without requiring retraining or fine-tuning for each new distribution.

    \item Extensive quantitative experiments demonstrate superior performance of \methodname{} compared to state-of-the-art unsupervised ($+17.95\%$ on PASCAL VOC 2012 \cite{pascalvoc2012} and $+5.20\%$ on MSCOCO-80 \cite{lin2014microsoft}) and weakly supervised semantic segmentation methods ($+21.67\%$ on PASCAL VOC 2012 \cite{pascalvoc2012} and $+7.99\%$ on MSCOCO-80 \cite{lin2014microsoft}) trained using Stable Diffusion \cite{rombach2022high}-generated synthetic images or web crawled images.
\end{itemize}
  
\section{Related Work}\label{sec:related_work}
\textbf{Efficient Annotation:} Efficient annotation methods aim to reduce the time and cost associated with the annotation process while maintaining or improving annotation quality. There are few works \cite{liao2021towards, branson2017lean} that use human-in-the-loop approaches to generate image-level annotation efficiently. Interactive object segmentation methods \cite{benenson2019large, ling2019fast} propose coarse segmentation masks which are then iteratively corrected by collecting point annotations (object vs background) from a human annotator. However, to the best of our knowledge, the research area for developing semantic segmentation annotation framework without access to any manual supervision remains unexplored.

\noindent
\textbf{Unsupervised Semantic Segmentation:} Unsupervised semantic segmentation \cite{cho2021picie, melas2022deep, vobecky2022drive, zhang2023rethinking, seong2023leveraging, zadaianchuk2022unsupervised, gao2022large} involves the segmentation of an image into meaningful regions without using labeled training data. Researchers have investigated diverse approaches to tackle this challenge, incorporating methods from unsupervised learning \cite{ke2022unsupervised, yu2023cross, tian2024diffuse}, clustering \cite{yu2022cmt, aflalo2023deepcut, huang2023style}, and self-supervised learning \cite{scheibenreif2022self, he2023clip, wang2023tokencut}. MaskContrast \cite{van2021unsupervised} stands out as an approach that performs unsupervised semantic segmentation within a contrastive learning framework. It assigns labels to the predicted clusters using Hungarian matching \cite{kuhn1955hungarian}, optimizing test-time performance metrics by associating labels with clusters that maximize segmentation accuracy. ACSeg \cite{li2023acseg} employs a pre-trained ViT model to mine ``concepts" from the pixel representation space of the unlabeled training images. Unlike previous methods, ACSeg  \cite{li2023acseg} does not predefine the number of clusters an image will be partitioned into. Instead, they adaptively conceptualize on different images due to varying complexity in individual images.

However, these existing unsupervised methods assume access to a large unlabeled training dataset. This contrasts with our problem formulation, where we assume a scenario without access to such unlabeled data. In our approach, we operate under the assumption that we lack prior knowledge of the specific data distribution the user requires for annotation. We perform on-the-fly inference with our \methodname{} model directly on the user's data, eliminating the need for pre-existing knowledge of the data distribution.

\noindent
\textbf{SAM Variants:} Several methods that build on SAM's capabilities have been developed recently. SemanticSAM \cite{li2023semantic} is a fully supervised method that segments and recognizes open-set objects at any granularity, and is trained on a combination of the SA-1B dataset \cite{kirillov2023segment} with other panoptic and part-segmentation datasets. In contrast, we neither have access to in-distribution training data, nor pixel-level annotations or image-level object labels. Our only requirement is the user-defined dataset-level object labels, meaning we lack information on which objects are present in each image. RegionSpot \cite{yang2023recognize} leverages the SAM model and CLIP's \cite{radford2021learning} Vision-Language feature embeddings to perform object detection. However, the method is dependent on (unlabeled) box annotations for every image, necessitating a comprehensive annotation effort for each object category. Grounded-SAM \cite{GroundedSAM} and LangSAM \cite{LangSAM} are two similar open-source projects based on SAM which aim to segment and detect anything with natural language prompts. However, they are weakly supervised methods that require the object labels for every image, which are used to generate bounding box annotations using GroundingDINO \cite{liu2023grounding}, which in turn serves as the box prompts for SAM. In our work, we do not utilize such object-level annotations, which may be difficult to get in many applications.

\begin{figure*}[t]
    \centering
    \includegraphics[width=\textwidth]{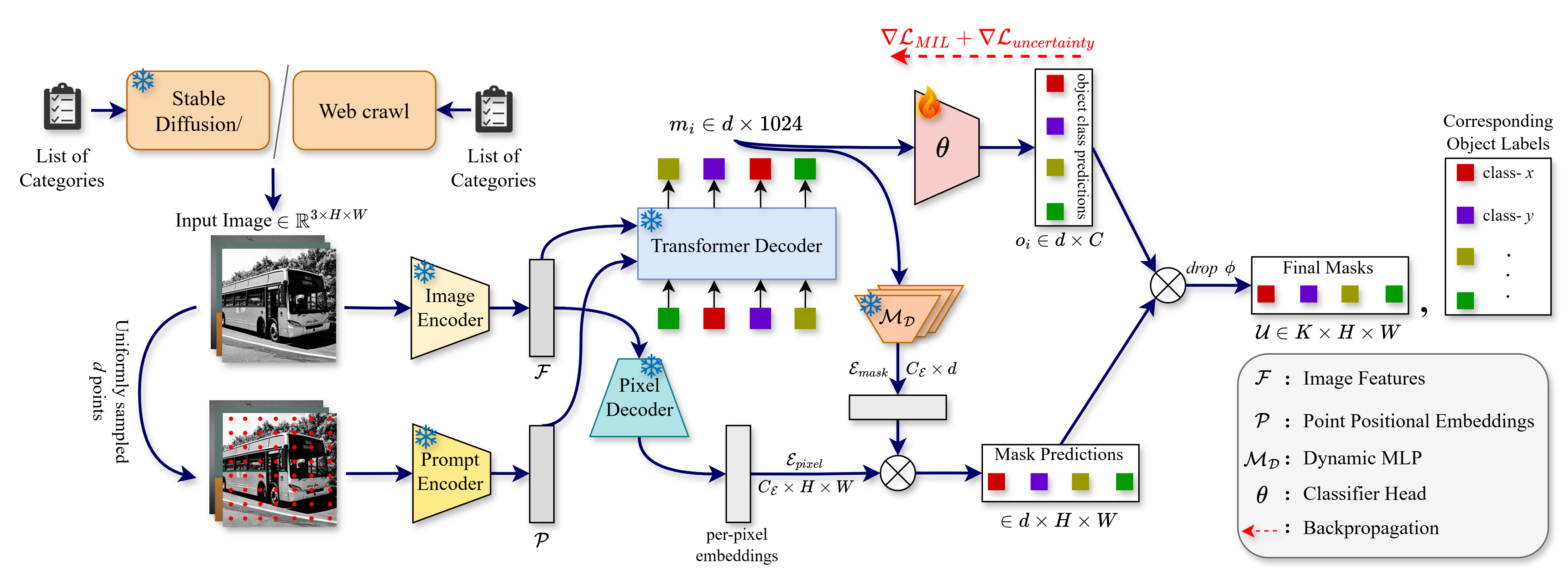}
    \vspace{-1em}
    \caption{\textbf{Overview of the \methodname{} architecture:} Given the list of user-defined target categories $\mathcal{C}$, we use Stable Diffusion \cite{rombach2022high} to generate a synthetic single-object image dataset which is encoded by SAM's \cite{kirillov2023segment} image encoder and a uniformly spaced grid of $d$ points are generated across the image to prompt SAM. The image and point embeddings are passed into a transformer decoder, the output mask embeddings $m_i$ (here, $m_i\in \mathbb{R}^{d\times 1024}$, corresponding to the $d$ masks predicted by SAM) of which are used to train a classifier head $\theta$ to predict objects using a Multiple Instance Learning (MIL) setup and uncertainty losses (refer Sec. \ref{subsec:classifier_head}).}
    \vspace{-1em}
    \label{fig:overall}
\end{figure*}  
\section{Methodology}
\label{sec:methodology}
\noindent
In this section, we define the problem statement (Sec. \ref{subsec:problem_formulation}) and present an overview of our \methodname{} framework (Sec. \ref{subsec:framework_overview}). We then briefly explain SAM's architecture (Sec. \ref{subsec:sam_arch}) and describe our strategy for collecting semantic information on different classes of interest (Sec. \ref{subsec:StableDiffusion}). Finally, we explain how this semantic information is linked with SAM's mask embeddings to develop our annotation system (Sec. \ref{subsec:classifier_head}) and outline the inference steps of the annotation system (Sec. \ref{subsec:Inference}).

\subsection{Problem Statement}\label{subsec:problem_formulation}
\noindent
Consider we have a set of unlabeled $N$ images $\mathcal{X} = \{x_i\}^N_{i=1}$ where, $x_i \in \mathbb{R}^{H\times W \times 3}$. Given a set of $C+1$ target classes $\mathcal{C} = \{c_i\}^C_{i=0}$ ($C$ object classes and one background class), our objective is to generate semantic masks $\mathcal{U}=\{u_i\}^N_{i=1}$ for the unlabeled $N$ images. Here, $u_i \in \mathcal{C}^{H\times W}$ and each image can contain variable number of objects. We also consider that apart from $\mathcal{X}$, we do not have access to more unlabeled images from the same distribution for unsupervised/self-supervised training.

\subsection{Framework Overview}\label{subsec:framework_overview}
\noindent
To generate semantic segmentation masks, the system needs to be able to distinguish different regions of an image and understand the semantics of those regions simultaneously. We leverage domain agnostic Segment Anything Model (SAM) \cite{kirillov2023segment} as the backbone of the system to distinguish the regions. Although SAM can generate masks at any granularity based on input prompts, it does not have the semantic understanding of the objects present in an image. To guide the semantic understanding, we utilize Stable Diffusion generated synthetic images/web crawled images of our classes of interest. Since, we know the image-level labels of the synthetic or web crawled images, we train a classifier head on top of SAM's mask decoder in a weakly-supervised setup. We learn a mapping for every generated mask of SAM by parsing the mask embedding and employing Multi-Layer Perceptron (MLP) layers to map it to class labels. We use multiple instance learning loss to learn the mapping function. Since SAM mask embeddings are domain agnostic, our learned classifier is also robust to distribution shifts and can be directly applicable to the unlabeled dataset of interest. Figure \ref{fig:overall} provides a detailed architectural overview.

\subsection{Segment Anything Model Architecture}\label{subsec:sam_arch}
\noindent
Segment Anything Model (SAM) \cite{kirillov2023segment} is a foundation model for image segmentation task trained on a dataset consisting of 11M images and 1B masks. It can generate segmentation masks for images at varying granularity levels based on input prompts. It consists of three main parts:

\noindent
\textbf{Image Encoder:} The input images are processed using a pre-trained Vision Transformer \cite{dosovitskiy2020image} modified to handle high resolution images of dimension $\mathbb{R}^{3\times 1024 \times 1024}$. It is run once per image to obtain features of dimension $\mathbb{R}^{256\times 64 \times 64}$ and can be applied prior to prompting the model.

\noindent
\textbf{Prompt Encoder:} SAM takes sparse prompt inputs in the form of point or box annotations for mask inference. Alternatively, in SAM's ``\textit{automatic}" mode, users don't need to manually define points or bounding boxes in the image; instead, a grid of uniformly spaced points (say, `$d$' number of total points) is generated throughout the image which are used as point annotations to prompt SAM. These point prompts are represented as positional embeddings summed with the learned embeddings (that indicates if a point is in the foreground or background) in SAM. Since, we do not have any prior information on where the object of interest is located, we use this ``\textit{automatic}" mode to generate all possible masks.

\noindent
\textbf{Mask Decoder:} SAM's mask decoder consists of a modified Transformer Decoder block \cite{vaswani2017attention} followed by a dynamic mask prediction heads. It maps the image and prompt embeddings along with an output token to a mask. A learned output token embedding which is analogous to the \verb|[class]| token in \cite{dosovitskiy2020image}, is first inserted into the set of prompt embeddings. This, along with the image embeddings are given as input to the transformer decoder that updates the output token embeddings (denoted by $m_{tokens}\in \mathbb{R}^{d\times 4 \times256}$). The image embeddings are upscaled by 4$\times$ and a point-wise product is taken with the output of a 3-layer MLP whose inputs are $m_{tokens}$. This product is the mask prediction of SAM, and this is performed for all of the $d$ prompts (point annotations) resulting in a set of $d$ binary masks $\in \mathbb{Z}_2^{d \times 256 \times 256}$ for each image. 



\subsection{Fetching Semantic Information}\label{subsec:StableDiffusion}
\noindent
Although SAM \cite{kirillov2023segment} can predict masks by distinguishing different regions of the image, it does not have semantic understanding of the predicted masks. To enable SAM with recognition capability of user defined classes, we collect semantic information of those classes using synthetic images or web crawled images.

\noindent
\textbf{Synthetic Images:} We use the Stable Diffusion \cite{rombach2022high} model which can generate hyper-realistic images given a text prompt. For each of the user-defined object classes in $\mathcal{C}$, we use the text prompt ``a photo of a \{c\}" $\forall c \in \mathcal{C}$ and generate 200 single object images. This synthetic single-object image dataset serves as guide to enable semantic understanding for Segment Anything Model.

\noindent
\textbf{Web Crawled Images:} Alternative to synthetic images, we can also use web crawled images for the same purpose. Through web crawling, we acquire single-object images for each user-defined object class in $\mathcal{C}$ by utilizing them as search queries on the web. The web crawled images serve as a valuable addition to the training data, providing diversity in backgrounds and object contexts. This approach enhances the model's ability to generalize across various visual scenarios and improves its discriminatory capacity for user-defined object classes.

\subsection{Enabling Semantic Recognition}\label{subsec:classifier_head}
\noindent
The $d$ masks generated by SAM for each image contains semantic maps at all granularity and also includes background. Our objective is to identify mask(s) containing the desired object(s) specified by the user.

We introduce a classifier head, denoted as $\theta$, to process SAM's flattened mask embeddings $m \in \mathbb{R}^{d\times 1024}$. The classifier head $\theta$ is a simple 4-layer MLP network. Its output $o_i = \theta(m_i)$, where $o_i \in \mathbb{R}^{d\times C}$, corresponds to the assigned object labels. In training $\theta$ with synthetic single-object data, each image has a 1-dimensional object label within a $d$-dimensional input space ($d$ mask embeddings per image). This setup necessitates a weakly supervised learning approach, leading us to adopt multiple-instance learning.

\noindent
\textbf{Multiple Instance Learning (MIL):} In MIL, the $d$ samples (SAM's mask embeddings $m_i$ for the $i^{th}$ image) are grouped into `positive' and `negative' bags. `Positive' bags contain at least one positive instance (a mask embedding $m_i^j \in \mathbb{R}^{1\times 1024}$ corresponding to the ground truth object label $y_i$), while `negative' bags have no positive instances. The objective is to train a model that can classify the entire bag and distinguish each instance within it.

To calculate the loss for each bag, denoted as the classifier output `$o_i$' in our problem, we aim to express each $d$-dimensional mask embedding through a single confidence score for each category. For a set of mask embeddings associated with a given image, the activation score for a specific category is computed as the average of the $k$-max activations across the number of masks ($d$) for that category. In our case, the dimension $d=100$ remains constant for all inputs, and we set $k = \max\left(1, \left\lceil\frac{d}{a}\right\rceil \right)$, where $a$ is a design parameter. Thus, our class-wise confidence scores for the $j^{th}$ class of the $i^{th}$ input can be represented as,

\begin{equation}
    s^j_i = \frac{1}{k} \max_{\substack{\mathcal{O}\subset o_i[:, j] \\ |\mathcal{O}|=k}}\sum_{l=1}^{k} \mathcal{O}_l .
\end{equation}

Subsequently, a softmax non-linearity is employed to derive the probability mass function across all categories as, $p^j_i = \frac{\exp(s^j_i)}{\sum_{j=1}^{C} \exp(s_j^i)}$. It is necessary to compare this probability mass function with the actual distribution of labels for each image to calculate the MIL loss. Since each image (during test-time) can have multiple objects in it, we represent the labels as a multi-hot vector, where 1 occurs if the object is present, else 0. We then normalize the ground truth label vector transforming it into a valid pmf. The MIL loss is subsequently computed as the cross-entropy between the predicted pmf, denoted as $\mathbf{p_i}$, and the normalized ground truth ($y_i = \left[ 
y_i^1,\hdots,y_i^C \right]^T$) for $N$ images and $C$ object categories as,

\vspace{-1em}
\begin{equation}
    \mathcal{L}_{MIL} = \frac{1}{N} \sum_{i=1}^{N} \sum_{j=1}^{C} -y_i^j \log(p_i^j).
\end{equation}

\noindent
\textbf{Uncertainty Distillation:} To further improve the discriminative ability of the object classifier, we employ an Uncertainty Distillation setup by leveraging the uncertainty information during training. The MIL trained classifier from the previous stage acts as the teacher model $\theta_t$ (with parameters frozen), and the student model $\theta_s$ is an untrained classifier having the same architecture as $\theta_t$. The primary motivation behind uncertainty distillation lies in enhancing the model's ability to make well-calibrated predictions and acknowledge its own uncertainties in challenging scenarios.

For the mask embeddings of the $i^{th}$ image, $m_i$, we compute the teacher logits as $o^t_i = \theta_t(m_i)$ and the teacher predictions as $\hat{y}^t_i = argmax(p^t_i)$, where $p^t_i = softmax(o^t_i) \in \mathbb{R}^{d\times C}$ are the object-label prediction probability scores for all the $d$ masks. We calculate the entropy (uncertainty) in the teacher logits as $H^t_i = - \sum_{j=1}^{C} p^t_{ij} \cdot \log(p^t_{ij})$. The goal is to minimize the reliance on (1) high entropy predictions, denoted by $H^{t, high}_i$, which are obtained by thresholding $H^t_i$ (such that,  $H^{t, high}_i = H^t_i>threshold$ and $H^{t, low}_i = H^t_i<threshold$), and (2) low entropy \textbf{\textit{incorrect}} predictions, denoted as $H^{t, low, incorr}_i = H^{t, low}_i \cap \{\hat{y}^t_i \neq y_i\}$, where $y_i$ is the ground truth object label. Thus, we obtain a set of bad predictions as $B_i = H^{t, high}_i \cup H^{t, low, incorr}_i$. Here, $B_i$ represents the indices of the masks out of all the $d$ masks predicted by SAM for the $i^{th}$ image, where the classifier head predicts a label with highly uncertainty, or predicts a wrong label confidently.

Next, we obtain the student logits as $o^s_i = \theta_s(m_i)$ and compute the MIL loss, $\mathcal{L}_{MIL}$, as per the previous section. With the obtained bad teacher prediction indices, we compute the bad student logits as $o^{s, bad}_i = o^s_i[B_i,:]$. Finally, we compute the uncertainty loss (to be minimized) as,
\begin{equation}
    \mathcal{L}_{uncertainty} = \frac{-1}{N} \sum_{i=1}^N \sum_{j=1}^C \log \left( \frac{\exp(o^{s, bad}_{ij})}{\sum_{j=1}^C\exp(o^{s, bad}_{ij})} \right).
\end{equation}

\noindent
\textbf{Overall Objective:} The net loss used to guide the student model is a weighted sum of the MIL and uncertainty losses as, $\mathcal{L}_s = \lambda_1 \cdot \mathcal{L}_{MIL} + \lambda_2 \cdot \mathcal{L}_{uncertainty}$. The trained student model $\theta_s$ is used for inference on the test data. 

\subsection{Inference}\label{subsec:Inference}
\noindent
For the images in the test set, we obtain the mask embeddings $m_i$ from SAM in the same way as during training on synthetic / web crawled images. These embeddings are passed through our trained classifier head $\theta_s$ to acquire object-class confidence scores for each mask within an image. By applying a predefined threshold to the confidence scores, we selectively retain masks whose object labels have been predicted with high probability. Among the selected subset masks for each image, certain masks may be redundant. We use Non-Maximal Suppression (NMS) to eliminate those overlapping masks and ensure the retention of masks only corresponding to distinct objects. 
  




\section{Experiments}\label{sec:experiment}
SAM's automatic mask generation mode uniformly places a grid of $d$ points across the image, and for \methodname{}, we set $d=100$. We extracted and saved all $d$ predicted masks from SAM, with the mask embeddings ($m_i \in \mathbb{R}^{d\times 1024}$). This one-pass extraction process ensures efficient storage for subsequent use in our workflow. We generated 200 synthetic single object images per class using the StableDiffuson-v1-4 \cite{rombach2022high} model. We first train the classifier head (teacher) $\theta_t$, with the MIL loss (${L}_{MIL}$) as the sole objective with batch size $64$, learning rate $0.001$ optimized with the Adam optimizer. With these same hyperparameters, we train the student model $\theta_s$ with the objective $\mathcal{L}_s = \lambda_1\cdot \mathcal{L}_{MIL} +  \lambda_2\cdot \mathcal{L}_{uncertainty}$, where we set $\lambda_1 = 1$ and $\lambda_2 = 0.15$. Our quantitative evaluation of \methodname{} relies on two metrics: mean Intersection over Union (mIoU) and mean Average Precision at 50\% IoU threshold (mAP$_{50}$). For our entire workflow, we utilized a single NVIDIA GeForce RTX 3090 GPU. We perform our experiments on two popular benchmarks: PASCAL VOC \cite{pascalvoc2012} and the MSCOCO-80 \cite{lin2014microsoft} datasets.




\begin{table}
\centering
\caption{\textbf{SOTA Comparison on PASCAL VOC} \textit{val} data. \methodname{} significantly outperforms state-of-the-art methods like TransFGU \cite{yin2022transfgu} and Leopart \cite{ziegler2022self} (unsupervised) and ACR \cite{kweon2023weakly} (weakly supervised) when these methods are trained on synthetic or web crawled images. This underscores \methodname{}’s robustness in scenarios lacking in-distribution training data. \textit{US} and \textit{WSS} denote Unsupervised and Weakly-Supervised Segmentation, respectively.
}
\vspace{-1em}
\resizebox{\columnwidth}{!}{
\begin{tabular}{c|cc|c|c}
\hline
\multirow{2}{*}{\textbf{Method}}                 & \multicolumn{2}{c|}{\textbf{Training Data}}                                           & \multirow{2}{*}{\textbf{mIoU}}                               & \multirow{2}{*}{\textbf{mAP$_{50}$}} \\ \cline{2-3}
                                                 & \multicolumn{1}{c|}{\textbf{Synthetic}}         & \textbf{Web Crawl}                  &                                                              &                                    \\ \hline
Leopart \cite{ziegler2022self} \textit{(US)}  & \multicolumn{1}{c|}{\ding{51}} &                                     & \underline{7.21\%}                                                       & -                                  \\ 
TransFGU \cite{yin2022transfgu} \textit{(US)} & \multicolumn{1}{c|}{\ding{51}} &                                     & 2.05\%                                                       & -                                  \\ 
ACR \cite{kweon2023weakly} \textit{(WSS)}      & \multicolumn{1}{c|}{\ding{51}} &                                     & 3.49\%                                                       & -                                  \\ \arrayrulecolor[gray]{0.85} \hline \arrayrulecolor{black}
\textbf{\methodname{}}          & \multicolumn{1}{c|}{\ding{51}} &                                     & \begin{tabular}[c]{@{}c@{}}\textbf{25.16\%} \\ {\color[HTML]{009901}{(+17.95\%)}}\end{tabular} & \textbf{49.06\%}                   \\ \hline
Leopart \cite{ziegler2022self} \textit{(US)}  & \multicolumn{1}{c|}{}                           & \ding{51}          & 6.79\%                                                       & -                                  \\ 
TransFGU \cite{yin2022transfgu} \textit{(US)} & \multicolumn{1}{c|}{}                           & \ding{51}          & 2.45\%                                                       & -                                  \\ 
ACR \cite{kweon2023weakly} \textit{(WSS)}      & \multicolumn{1}{c|}{}                           & \ding{51}          & \underline{13.42\%}                                                      & -                                  \\ \arrayrulecolor[gray]{0.85} \hline \arrayrulecolor{black}
\textbf{\methodname{}}          & \multicolumn{1}{c|}{}                           & \textbf{\ding{51}} & \begin{tabular}[c]{@{}c@{}}\textbf{22.42\%} \\ {\color[HTML]{009901}{(+9.00\%)}}\end{tabular}  & \textbf{45.59\%}                   \\ \hline
\end{tabular}
}
\label{tab:comparison_pascal}
\end{table}

\begin{table}
\centering
\caption{\textbf{SOTA Comparison on MSCOCO-80} \textit{val} data. We observe that \methodname{} significantly outperforms state-of-the-art methods like TransFGU \cite{yin2022transfgu} and Leopart \cite{ziegler2022self} (unsupervised) and ACR \cite{kweon2023weakly} (weakly supervised) when trained on synthetic or web crawled images. This highlights the reliance of these methods on having the train and test data from the same distribution. Notably, they struggle when the train and test distributions mismatch, even with identical class spaces. \textit{US} and \textit{WSS} denote Unsupervised and Weakly-Supervised Segmentation, respectively.}
\vspace{-1em}
\resizebox{\columnwidth}{!}{
\begin{tabular}{c|cc|c|c}
\hline
\multirow{2}{*}{\textbf{Method}}                      & \multicolumn{2}{c|}{\textbf{Training Data}}                                           & \multirow{2}{*}{\textbf{mIoU}}                             & \multirow{2}{*}{\textbf{mAP$_{50}$}} \\ \cline{2-3}
                                                      & \multicolumn{1}{c|}{\textbf{Synthetic}}         & \textbf{Web Crawl}                  &                                                            &                                    \\ \hline
Leopart \cite{ziegler2022self} (US)  & \multicolumn{1}{c|}{\ding{51}} &                                     & \underline{3.84}\%                                                     & -                                  \\
TransFGU \cite{yin2022transfgu} (US) & \multicolumn{1}{c|}{\ding{51}} &                                     & 0.95\%                                                     & -                                  \\
ACR \cite{kweon2023weakly} (WSS)     & \multicolumn{1}{c|}{\ding{51}} &                                     & 0.61\%                                                     & -                                  \\ \arrayrulecolor[gray]{0.85} \hline \arrayrulecolor{black}
\textbf{\methodname{}}               & \multicolumn{1}{c|}{\ding{51}} &                                     & \begin{tabular}[c]{@{}c@{}}\textbf{8.60\%}\\ {\color[HTML]{009901}{(+4.76\%)}}\end{tabular} & \textbf{30.90\%}                   \\ \hline
Leopart \cite{ziegler2022self} (US)  & \multicolumn{1}{c|}{}                           & \ding{51}          & \underline{3.81}\%                                                     & -                                  \\
TransFGU \cite{yin2022transfgu} (US) & \multicolumn{1}{c|}{}                           & \ding{51}          & 1.02\%                                                     & -                                  \\
ACR \cite{kweon2023weakly} (WSS)     & \multicolumn{1}{c|}{}                           & \ding{51}          & 2.10\%                                                     & -                                  \\ \arrayrulecolor[gray]{0.85} \hline \arrayrulecolor{black}
\textbf{\methodname{}}               & \multicolumn{1}{c|}{}                           & \textbf{\ding{51}} & \begin{tabular}[c]{@{}c@{}}\textbf{9.01\%}\\ {\color[HTML]{009901}{(+5.20\%)}}\end{tabular} & \textbf{31.80\%}                   \\ \hline
\end{tabular}
}
\vspace{-2em}
\label{tab:comparison_coco}
\end{table}

\noindent
\textbf{Quantitative Results:} \methodname{} is trained on synthetic/web crawled data to recognize object semantics and applied for inference on the \textit{val} sets of PASCAL VOC and MSCOCO-80 datasets. While unsupervised semantic segmentation methods are a common choice for annotation without manual supervision, state-of-the-art (SOTA) unsupervised methods typically rely on large-scale unlabeled datasets from the same distribution as the test set \cite{cho2021picie, yin2022transfgu, ziegler2022self}. Given our problem setup that we lack access to such large-scale unlabeled data from the same distribution, a fair comparison with existing SOTA unsupervised methods is challenging. To establish a fair baseline for comparison, we setup two unsupervised methods, Leopart \cite{ziegler2022self} and TransFGU \cite{yin2022transfgu}, and one weakly supervised method, ACR \cite{kweon2023weakly}, trained on our synthetic/web crawled data and evaluated on the PASCAL VOC and MSCOCO-80 datasets. In Tables \ref{tab:comparison_pascal} and \ref{tab:comparison_coco}, our focus is on the performance of Leopart \cite{ziegler2022self}, TransFGU \cite{yin2022transfgu} and ACR \cite{kweon2023weakly} (which achieved high results on in-domain evaluations) when trained on our synthetic/web crawled data. \textit{This demonstrates that these existing methods do not work well in unseen domains and require access to data from the same distribution as the test set in order to train or adapt their models.} \methodname{} does not have this limitation, and thus could be applied across a large variety of applications.

Tables \ref{tab:comparison_pascal} and \ref{tab:comparison_coco} present performance comparisons between \methodname{} and SOTA trained on synthetic or web crawled data and evaluated on the PASCAL VOC and MSCOCO-80 datasets, respectively. On PASCAL VOC, \methodname{} demonstrates competitive performance compared to unsupervised methods like Leopart \cite{ziegler2022self} and TransFGU \cite{yin2022transfgu}, as well as the weakly supervised method ACR \cite{kweon2023weakly}, when these methods are trained on synthetic/web crawled data. This highlights that these SOTA methods are heavily reliant on the test data belonging to the same distribution as the training data, a requirement that \methodname{} does not have, showcasing its robustness across diverse datasets.

On the MSCOCO-80 dataset, which features diverse object classes and complex scenes, all methods face significant challenges. Despite these challenges, \methodname{} maintains a competitive edge when compared to Leopart \cite{ziegler2022self}, TransFGU \cite{yin2022transfgu}, and ACR \cite{kweon2023weakly} trained on synthetic/web crawled data. This underscores that these methods are heavily dependent on the alignment between the train and test data distributions, a limitation that \methodname{} avoids. The results on MSCOCO-80, as shown in Table \ref{tab:comparison_coco}, reveal a trade-off between localization precision and segmentation accuracy. A high mAP$_{50}$ score indicates proficiency in object detection and localization, while a low mIoU score suggests difficulties in precisely delineating object boundaries or capturing fine-grained details.

\noindent
\textbf{Discussion:} It is essential to highlight a key aspect of our methodology: our training data originates from a small set of synthetic ($200$ images per class) or web crawled data ($\sim45$ images per class) which are single-object images, and thus distinctly different from the test data distribution. Unlike existing methods, which often rely on detailed annotations for each image, \emph{our method only requires knowledge of the user-defined class names to be segmented across the entire dataset}. The ability to apply an existing method to a new dataset, as achieved by \methodname{}, without any knowledge of the distribution of the data in this dataset is a highly desired feature as it enhances generalizability well beyond what unsupervised domain adaptation can do. 

Furthermore, the effectiveness of our method is evident in its ability to yield competitive results regardless of the data source, be it synthetic or web crawled images. When TransFGU \cite{yin2022transfgu}, Leopart \cite{ziegler2022self} and ACR \cite{kweon2023weakly} are trained on our synthetic or web crawled datasets, the performance drops significantly compared to their performance on in-distribution training data in both PASCAL VOC and MSCOCO-80 datasets, emphasizing the unique challenges posed by our problem setting. On the other hand, \methodname{} achieves $+17.95$\% and $+5.17$\% mIoU, compared to Leopart \cite{ziegler2022self}. This flexibility underscores the robustness of our approach, allowing for practical applications in scenarios where obtaining extensive manual annotations per image is challenging or impractical.

\begin{table}[]
\caption{Comparison of \methodname{} performance with the SAM+CLIP \cite{radford2021learning} method on the PASCAL VOC and MSCOCO-80 datasets. \textit{SD}: Training set is Stable Diffusion generated synthetic images; \textit{WC}: Training set is web crawled images.}
\vspace{-1em}
\label{tab:clip_comparison}
\centering
\resizebox{\columnwidth}{!}{
\begin{tabular}{lcc|lcc}
\hline
\multicolumn{3}{c|}{\textbf{PASCAL VOC}}                                                                      & \multicolumn{3}{c}{\textbf{MSCOCO-80}}                                                                        \\ \hline
\multicolumn{1}{l|}{\textbf{Method}}                                & \multicolumn{1}{c|}{\textbf{mIoU}}    & \textbf{mAP}$\mathbf{_{50}}$ & \multicolumn{1}{l|}{\textbf{Method}}                              & \multicolumn{1}{c|}{\textbf{mIoU}}   & \textbf{mAP}$\mathbf{_{50}}$ \\ \hline
\multicolumn{1}{l|}{SAM+CLIP}                                  & \multicolumn{1}{c|}{19.17\%} & 42.45\%    & \multicolumn{1}{l|}{SAM+CLIP}                                & \multicolumn{1}{c|}{7.91\%} & 29.93\%    \\
\multicolumn{1}{l|}{\textbf{\methodname{} \textit{(SD)}}} & \multicolumn{1}{c|}{\textbf{25.16\%}} & \textbf{49.06\%}    & \multicolumn{1}{l|}{\methodname{} \textit{(SD)}} & \multicolumn{1}{c|}{8.60\%} & 30.90\%    \\
\multicolumn{1}{l|}{\methodname{} \textit{(WC)}} & \multicolumn{1}{c|}{22.42\%} & 45.59\%    & \multicolumn{1}{l|}{\textbf{\methodname{} \textit{(WC)}}} & \multicolumn{1}{c|}{\textbf{9.01\%}} & \textbf{31.80\%}    \\ \hline
\end{tabular}
}
\vspace{-2em}
\end{table}

\begin{figure*}
    \centering
    \includegraphics[width=\textwidth]{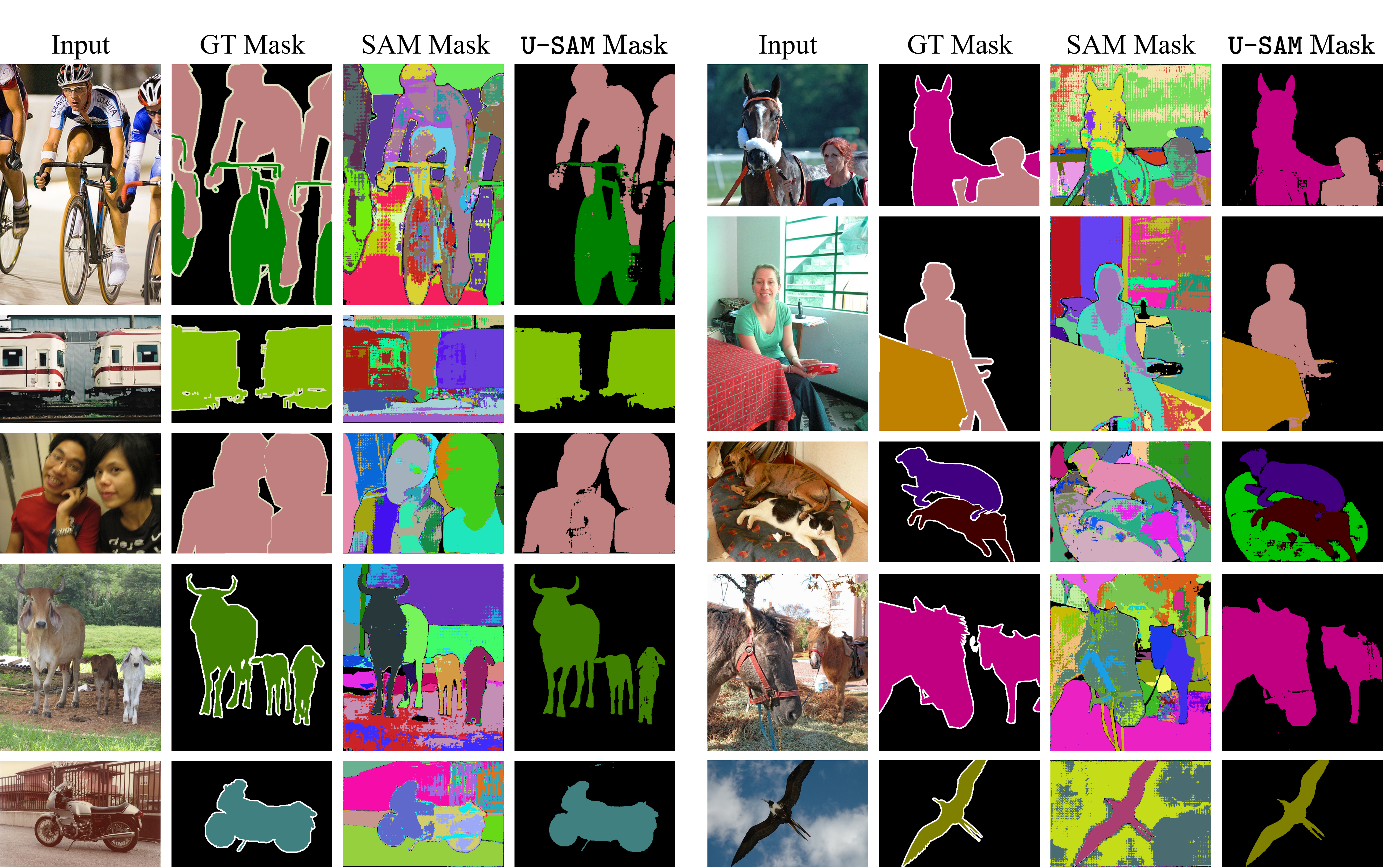}
    \vspace{-2em}
    \caption{\textbf{Qualitative results:} The original image, the ground truth (GT) mask, all of the SAM generated masks overlayed on top of each other, and \methodname{} predicted masks are shown respectively in the four columns. For the GT mask and \methodname{} predicted masks, the colors indicate the class labels, while random colors were used for the SAM masks since it does not provide class-labels.}
    \vspace{-1em}
    \label{fig:qualitative_pascal}
\end{figure*}

\noindent
\textbf{Additional baseline:} We incorporated the CLIP \cite{radford2021learning} model, renowned for its zero-shot performance, as an additional baseline to compare \methodname{}'s performance. The baseline setup for CLIP involves computing the similarity between images and a set of natural language words/phrases (object labels in our case), leveraging its shared Vision-Language embedding space. Utilizing the $d$ masks predicted by SAM for each image, we generate $d$ pooled images, by multiplying the masks to the original image, with each corresponding to a distinct object category. Subsequently, we calculate the similarity between all $d$ pooled images and each of the $C$ user-defined object classes. Similar to our \methodname{} post-processing, we set a similarity threshold at 70\% and apply Non-Maximal Suppression to retain only unique objects while eliminating redundant masks. The results obtained by this SAM+CLIP \cite{radford2021learning} method compared to our \methodname{} model are reported in Table \ref{tab:clip_comparison}. Unlike SAM+CLIP \cite{radford2021learning}, which lacks granularity matching and relies on CLIP's training data, \methodname{} uses web-crawled data, making it more versatile across diverse datasets like PASCAL VOC \cite{pascalvoc2012} and MSCOCO-80 \cite{lin2014microsoft}.

\noindent
\textbf{Qualitative Results:} In Figure \ref{fig:qualitative_pascal}, we showcase qualitative results obtained by \methodname{}. In each example, SAM generates multiple masks at various granularities, including background masks. For instance, in the first example featuring a person on a bicycle, SAM produces separate masks for the bike helmet, the person's arms, and legs. In contrast, \methodname{} provides masks only for the objects requested by the user, demonstrating targeted segmentation.

The second example illustrates two disconnected objects of the same class, while the third example shows two objects of the same class with a connected contour. In all cases, \methodname{} successfully filters out the desired mask with the correct object label. Unlike SAM, which segments numerous parts of the image—both small and large objects—\methodname{} focuses on the user-specified classes. For example, in the third example (left set), SAM segments the eyes and face separately, but \methodname{} outputs only the person mask, as it aligns with the user-defined class. This demonstrates that \methodname{} achieves the desired granularity level.

Furthermore, in the third example of the right set in Figure \ref{fig:qualitative_pascal}, the ground truth mask lacks annotation for the ``sofa" object. However, \methodname{} correctly identifies and outputs the sofa mask, as it falls within the user-requested object class space. This highlights \methodname{}’s ability to provide accurate and relevant segmentation masks, which might have even been missed by the ground truth masks.

\noindent
\textbf{Changed Granularity:} We modified the class definitions for the PASCAL VOC dataset to adjust the granularity of user-defined classes. Specifically, we introduced some superset classes that encompass several PASCAL VOC classes together. The revised class definitions and the corresponding PASCAL VOC classes falling into these categories are as follows:

\begin{enumerate}
    \item animals: ``\textit{bird}", ``\textit{cat}", ``\textit{cow}", ``\textit{dog}", ``\textit{horse}", ``\textit{sheep}"
    \item furniture: ``\textit{chair}", ``\textit{diningtable}", ``\textit{sofa}"
    \item household items: ``\textit{bottle}", ``\textit{pottedplant}", ``\textit{tvmonitor}"
    \item person: ``\textit{person}"
    \item transportation: ``\textit{aeroplane}", ``\textit{bicycle}", ``\textit{boat}", ``\textit{bus}", ``\textit{car}", ``\textit{motorbike}", ``\textit{train}"
\end{enumerate}

\begin{figure}
    \centering
    
    \includegraphics[width=\columnwidth]{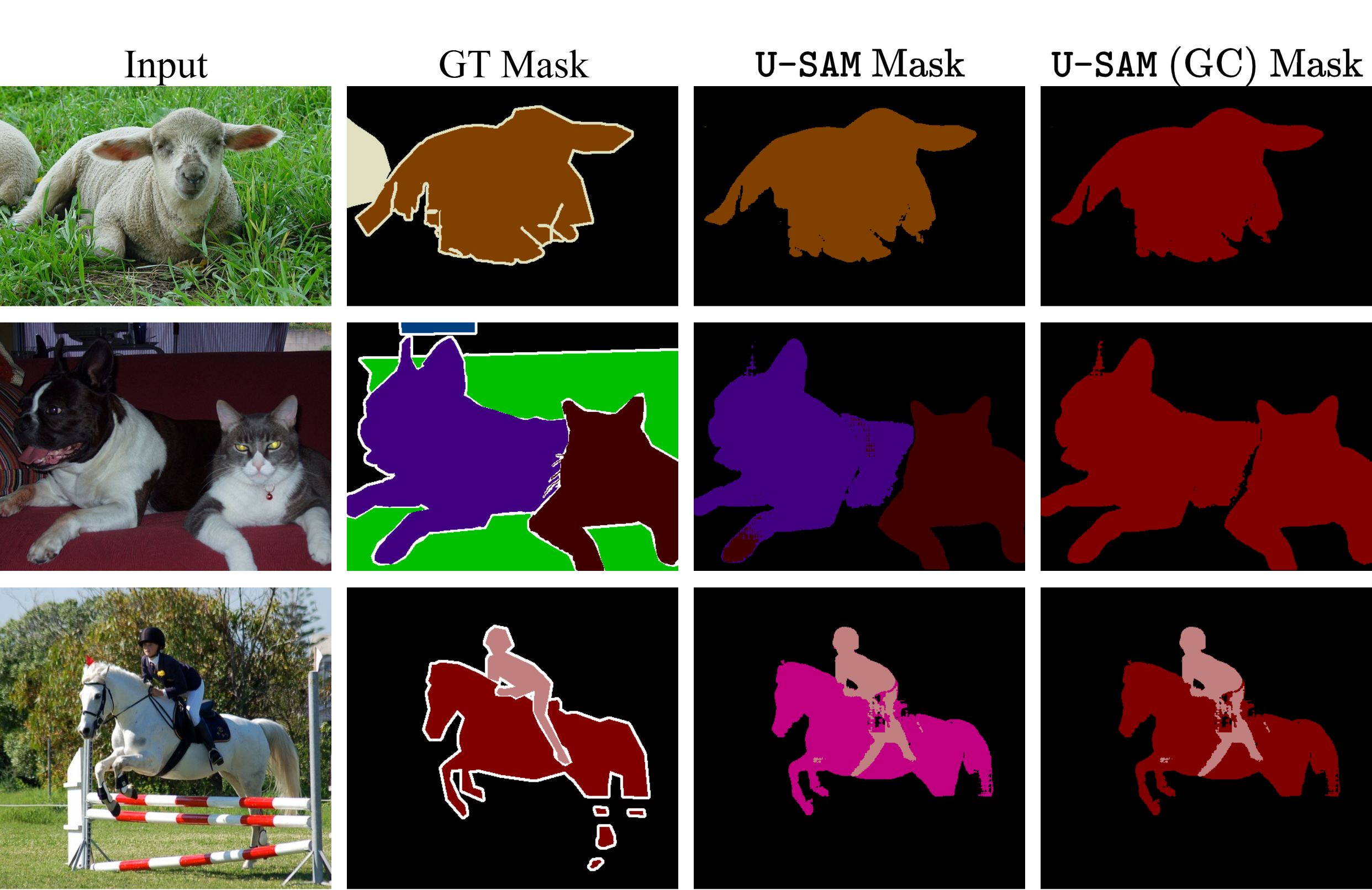}
    \vspace{-2em}
    \caption{\textbf{Qualitative results with changed granularity:} The third column shows the \methodname{} predictions on PASCAL classes, and the fourth column shows the predictions when the granularity level has been changed to categorize ``dog", ``cat" and ``sheep" classes as a single ``animals" class. The colors represent the different class labels: \textcolor{brown}{Brown}: ``sheep", \textcolor{violet}{Violet}: ``dog", \textcolor{olive}{Dark Brown}: ``cat", \textcolor{red}{Red}: ``animals", \textcolor{pink}{Pink}: ``person". \textit{GC}: Granularity Changed.}
    \label{fig:granularity}
    \vspace{-1em}
\end{figure}

We applied the \methodname{} model to the PASCAL VOC dataset with the changed granularities. For each of the five new classes, we generate 200 synthetic images using the Stable Diffusion model, extract SAM mask embeddings $m_i$ for each image, and train the classifier head $\theta$. This updated granularity configuration yielded an mAP$_{50}$ score of 61.54\%. Some of the visual results obtained are shown in Figure \ref{fig:granularity}. \methodname{} has been able to detect these classes in both granularity levels- one where the typical PASCAL VOC classes were used, and when the granularity level was changed by altering the definitions of the desired classes.

\begin{table}[t!]
\centering
\caption{\textbf{Ablation study results:} Uncertainty Distillation significantly improves performance in both synthetic and web crawled data training paradigms for the PASCAL VOC and MSCOCO-80 datasets. MIL: Multiple Instance Learning.}
\vspace{-1em}
\label{tab:ablation_loss}
\resizebox{\columnwidth}{!}{
\begin{tabular}{c|cc|c|c|c}
\hline
\multirow{2}{*}{\textbf{Method}} & \multicolumn{2}{c|}{\textbf{Training   Data}}                                & \multirow{2}{*}{\textbf{Test Data}} & \multirow{2}{*}{\textbf{mIoU}} & \multirow{2}{*}{\textbf{mAP}} \\ \cline{2-3}
                                 & \multicolumn{1}{c|}{\textbf{Synthetic}}                  & \textbf{Web crawled}                 &                                     &                                &                               \\ \hline
MIL only                         & \multicolumn{1}{c|}{\ding{51}} &                            & \multirow{4}{*}{PASCAL VOC}         & 22.42\%                        & 42.47\%                       \\ \arrayrulecolor[gray]{0.85} \cline{1-3} \cline{5-6} \arrayrulecolor{black} 
MIL +   Distillation             & \multicolumn{1}{c|}{\ding{51}} &                            &                                     & \begin{tabular}[c]{@{}c@{}}\textbf{25.16\%}\\ {\color[HTML]{009901}{(+2.53\%)}}\end{tabular}                        & \begin{tabular}[c]{@{}c@{}}\textbf{49.06\%}\\ {\color[HTML]{009901}{(+6.56\%)}}\end{tabular}                       \\ \cline{1-3} \cline{5-6} 
MIL only                         & \multicolumn{1}{l|}{}                           & \ding{51} &                                     & 17.42\%                        & 39.86\%                       \\ \arrayrulecolor[gray]{0.85} \cline{1-3} \cline{5-6} \arrayrulecolor{black} 
MIL +   Distillation             & \multicolumn{1}{l|}{}                           & \ding{51} &                                     & \begin{tabular}[c]{@{}c@{}}\textbf{22.42\%}\\ {\color[HTML]{009901}{(+5.00\%)}}\end{tabular}                        & \begin{tabular}[c]{@{}c@{}}\textbf{45.59\%}\\ {\color[HTML]{009901}{(+5.73\%)}}\end{tabular}                       \\ \hline
MIL only                         & \multicolumn{1}{c|}{\ding{51}} &                            & \multirow{4}{*}{MSCOCO-80}            & 8.01\%                         & 30.89\%                       \\ \arrayrulecolor[gray]{0.85} \cline{1-3} \cline{5-6} \arrayrulecolor{black} 
MIL +   Distillation             & \multicolumn{1}{c|}{\ding{51}} &                            &                                     & \begin{tabular}[c]{@{}c@{}}\textbf{8.60\%}\\ {\color[HTML]{009901}{(+0.59\%)}}\end{tabular}                & \begin{tabular}[c]{@{}c@{}}\textbf{30.97\%}\\ {\color[HTML]{009901}{(+0.08\%)}}\end{tabular}              \\ \cline{1-3} \cline{5-6} 
MIL only                         & \multicolumn{1}{l|}{}                           & \ding{51} &                                     & 8.57\%                         & 27.57\%                       \\ \arrayrulecolor[gray]{0.85} \cline{1-3} \cline{5-6} \arrayrulecolor{black} 
MIL +   Distillation             & \multicolumn{1}{l|}{}                           & \ding{51} &                                     & \begin{tabular}[c]{@{}c@{}}\textbf{9.01\%}\\ {\color[HTML]{009901}{(+0.44\%)}}\end{tabular}                & \begin{tabular}[c]{@{}c@{}}\textbf{31.81\%}\\ {\color[HTML]{009901}{(+4.24\%)}}\end{tabular}              \\ \hline
\end{tabular}
}
\vspace{-1.25em}
\end{table}

\noindent
\textbf{Ablation Study:} In our ablation study (Table \ref{tab:ablation_loss}), we investigated the impact of introducing uncertainty distillation alongside the MIL loss in our classifier head ($\theta$) training framework. The primary objective was to understand the performance implications of leveraging uncertainty information during training, particularly in challenging cases.

Our experiments yielded compelling results, showcasing significant performance improvements, particularly in the PASCAL VOC dataset (with a $+5\%$ increase in mIoU and $+6.59\%$ in mAP$_{50}$) when leveraging web crawled data for training. The strategic incorporation of Uncertainty Distillation alongside the MIL loss played a pivotal role in enhancing the model's discriminative capabilities. Notably, this improvement is pronounced in the case of MSCOCO-80, where the utilization of web crawled training data addresses the challenges posed by the dataset's diversity and complexity. The model's adept handling of prediction uncertainties proves crucial in navigating the intricacies of real-world scenarios presented by the MSCOCO-80 dataset.

A key observation is the model's improved robustness in scenarios where traditional methods might struggle, showcasing the practical relevance of incorporating Uncertainty Distillation. The ability to make well-calibrated predictions and acknowledge uncertainties translated into a more effective and reliable segmentation framework.

\section{Conclusion}\label{sec:conclusion}
\methodname{} represents a significant advancement in semantic segmentation by transforming the Segment Anything Model (SAM) into a semantically aware framework. SAM, while powerful in generating precise object masks, lacked the ability to recognize and label specific object categories. Our approach addresses this limitation by integrating a classifier head into SAM, trained using synthetic or web-crawled images based solely on user-provided object class names. This enhancement enables \methodname{} to generate targeted semantic masks for user-defined object categories without relying on in-domain training data or image-level class labels. Instead, our pipeline requires only dataset-level class names, making it highly efficient and flexible for real-world applications. By automating pixel-level annotations and eliminating the need for manual labeling, \methodname{} offers a reliable solution for semantic segmentation, capable of generalizing across diverse datasets without retraining or fine-tuning. This innovation paves the way for more accessible and adaptable semantic segmentation tools, aligning with the evolving needs of users who require precise object identification without the burden of extensive data preparation.  
\newpage
{
    \small
    \bibliographystyle{ieeenat_fullname}
    \bibliography{main}

\begin{thebibliography}{58}
\providecommand{\natexlab}[1]{#1}
\providecommand{\url}[1]{\texttt{#1}}
\expandafter\ifx\csname urlstyle\endcsname\relax
  \providecommand{\doi}[1]{doi: #1}\else
  \providecommand{\doi}{doi: \begingroup \urlstyle{rm}\Url}\fi

\bibitem[Aflalo et~al.(2023)Aflalo, Bagon, Kashti, and Eldar]{aflalo2023deepcut}
Amit Aflalo, Shai Bagon, Tamar Kashti, and Yonina Eldar.
\newblock Deepcut: Unsupervised segmentation using graph neural networks clustering.
\newblock In \emph{Proceedings of the IEEE/CVF International Conference on Computer Vision}, pages 32--41, 2023.

\bibitem[Basak et~al.(2022)Basak, Kundu, and Sarkar]{basak2022mfsnet}
Hritam Basak, Rohit Kundu, and Ram Sarkar.
\newblock Mfsnet: A multi focus segmentation network for skin lesion segmentation.
\newblock \emph{Pattern Recognition}, 128:\penalty0 108673, 2022.

\bibitem[Basak et~al.(2023)Basak, Chattopadhyay, Kundu, Nag, and Mallipeddi]{basak2023ideal}
Hritam Basak, Soumitri Chattopadhyay, Rohit Kundu, Sayan Nag, and Rammohan Mallipeddi.
\newblock Ideal: Improved dense local contrastive learning for semi-supervised medical image segmentation.
\newblock In \emph{ICASSP 2023-2023 IEEE International Conference on Acoustics, Speech and Signal Processing (ICASSP)}, pages 1--5. IEEE, 2023.

\bibitem[Benenson et~al.(2019)Benenson, Popov, and Ferrari]{benenson2019large}
Rodrigo Benenson, Stefan Popov, and Vittorio Ferrari.
\newblock Large-scale interactive object segmentation with human annotators.
\newblock In \emph{Proceedings of the IEEE/CVF conference on computer vision and pattern recognition}, pages 11700--11709, 2019.

\bibitem[Branson et~al.(2017)Branson, Van~Horn, and Perona]{branson2017lean}
Steve Branson, Grant Van~Horn, and Pietro Perona.
\newblock Lean crowdsourcing: Combining humans and machines in an online system.
\newblock In \emph{Proceedings of the IEEE Conference on Computer Vision and Pattern Recognition}, pages 7474--7483, 2017.

\bibitem[Caron et~al.(2020)Caron, Misra, Mairal, Goyal, Bojanowski, and Joulin]{caron2020unsupervised}
Mathilde Caron, Ishan Misra, Julien Mairal, Priya Goyal, Piotr Bojanowski, and Armand Joulin.
\newblock Unsupervised learning of visual features by contrasting cluster assignments.
\newblock \emph{Advances in neural information processing systems}, 33:\penalty0 9912--9924, 2020.

\bibitem[Chen et~al.(2023)Chen, Wang, Liu, He, Cong, and Wan]{chen2023edge}
Chen Chen, Chenyu Wang, Bin Liu, Ci He, Li Cong, and Shaohua Wan.
\newblock Edge intelligence empowered vehicle detection and image segmentation for autonomous vehicles.
\newblock \emph{IEEE Transactions on Intelligent Transportation Systems}, 24\penalty0 (11):\penalty0 13023--13034, 2023.

\bibitem[Cho et~al.(2021)Cho, Mall, Bala, and Hariharan]{cho2021picie}
Jang~Hyun Cho, Utkarsh Mall, Kavita Bala, and Bharath Hariharan.
\newblock Picie: Unsupervised semantic segmentation using invariance and equivariance in clustering.
\newblock In \emph{Proceedings of the IEEE/CVF Conference on Computer Vision and Pattern Recognition}, pages 16794--16804, 2021.

\bibitem[Dosovitskiy et~al.(2020)Dosovitskiy, Beyer, Kolesnikov, Weissenborn, Zhai, Unterthiner, Dehghani, Minderer, Heigold, Gelly, et~al.]{dosovitskiy2020image}
Alexey Dosovitskiy, Lucas Beyer, Alexander Kolesnikov, Dirk Weissenborn, Xiaohua Zhai, Thomas Unterthiner, Mostafa Dehghani, Matthias Minderer, Georg Heigold, Sylvain Gelly, et~al.
\newblock An image is worth 16x16 words: Transformers for image recognition at scale.
\newblock \emph{arXiv preprint arXiv:2010.11929}, 2020.

\bibitem[Everingham et~al.(2015)Everingham, Eslami, Van~Gool, Williams, Winn, and Zisserman]{pascalvoc2012}
Mark Everingham, SM~Ali Eslami, Luc Van~Gool, Christopher~KI Williams, John Winn, and Andrew Zisserman.
\newblock The pascal visual object classes challenge: A retrospective.
\newblock \emph{International journal of computer vision}, 111:\penalty0 98--136, 2015.

\bibitem[Gao et~al.(2022)Gao, Li, Yang, Cheng, Han, and Torr]{gao2022large}
Shanghua Gao, Zhong-Yu Li, Ming-Hsuan Yang, Ming-Ming Cheng, Junwei Han, and Philip Torr.
\newblock Large-scale unsupervised semantic segmentation.
\newblock \emph{IEEE transactions on pattern analysis and machine intelligence}, 2022.

\bibitem[Gong et~al.(2023)Gong, Yang, Yao, Chen, Wang, Du, He, and Liu]{gong2023integrated}
Liang Gong, Zhiyu Yang, Yihang Yao, Binhao Chen, Wenjie Wang, Xiaofeng Du, Yidong He, and Chengliang Liu.
\newblock An integrated in situ image acquisition and annotation scheme for instance segmentation models in open scenes with a human--robot interaction approach.
\newblock \emph{IEEE Transactions on Human-Machine Systems}, 2023.

\bibitem[Guo et~al.(2022)Guo, Lu, Hou, Liu, Cheng, and Hu]{guo2022segnext}
Meng-Hao Guo, Cheng-Ze Lu, Qibin Hou, Zhengning Liu, Ming-Ming Cheng, and Shi-Min Hu.
\newblock Segnext: Rethinking convolutional attention design for semantic segmentation.
\newblock \emph{Advances in Neural Information Processing Systems}, 35:\penalty0 1140--1156, 2022.

\bibitem[Guo et~al.(2021)Guo, Yang, Li, and Yuan]{guo2021metacorrection}
Xiaoqing Guo, Chen Yang, Baopu Li, and Yixuan Yuan.
\newblock Metacorrection: Domain-aware meta loss correction for unsupervised domain adaptation in semantic segmentation.
\newblock In \emph{Proceedings of the IEEE/CVF conference on computer vision and pattern recognition}, pages 3927--3936, 2021.

\bibitem[He et~al.(2023{\natexlab{a}})He, Li, Zhang, Xu, Tang, Zhang, Guo, and Li]{he2023weakly}
Chunming He, Kai Li, Yachao Zhang, Guoxia Xu, Longxiang Tang, Yulun Zhang, Zhenhua Guo, and Xiu Li.
\newblock Weakly-supervised concealed object segmentation with sam-based pseudo labeling and multi-scale feature grouping.
\newblock \emph{Advances in Neural Information Processing Systems}, 36:\penalty0 30726--30737, 2023{\natexlab{a}}.

\bibitem[He et~al.(2020)He, Fan, Wu, Xie, and Girshick]{he2020momentum}
Kaiming He, Haoqi Fan, Yuxin Wu, Saining Xie, and Ross Girshick.
\newblock Momentum contrast for unsupervised visual representation learning.
\newblock In \emph{Proceedings of the IEEE/CVF conference on computer vision and pattern recognition}, pages 9729--9738, 2020.

\bibitem[He et~al.(2023{\natexlab{b}})He, Jamonnak, Gou, and Ren]{he2023clip}
Wenbin He, Suphanut Jamonnak, Liang Gou, and Liu Ren.
\newblock Clip-s4: Language-guided self-supervised semantic segmentation.
\newblock In \emph{Proceedings of the IEEE/CVF conference on computer vision and pattern recognition}, pages 11207--11216, 2023{\natexlab{b}}.

\bibitem[Huang et~al.(2023)Huang, Chen, Li, Li, Li, Song, Yan, and Xiong]{huang2023style}
Wei Huang, Chang Chen, Yong Li, Jiacheng Li, Cheng Li, Fenglong Song, Youliang Yan, and Zhiwei Xiong.
\newblock Style projected clustering for domain generalized semantic segmentation.
\newblock In \emph{Proceedings of the IEEE/CVF conference on computer vision and pattern recognition}, pages 3061--3071, 2023.

\bibitem[Ke et~al.(2022)Ke, Hwang, Guo, Wang, and Yu]{ke2022unsupervised}
Tsung-Wei Ke, Jyh-Jing Hwang, Yunhui Guo, Xudong Wang, and Stella~X Yu.
\newblock Unsupervised hierarchical semantic segmentation with multiview cosegmentation and clustering transformers.
\newblock In \emph{Proceedings of the IEEE/CVF Conference on Computer Vision and Pattern Recognition}, pages 2571--2581, 2022.

\bibitem[Kirillov et~al.(2023)Kirillov, Mintun, Ravi, Mao, Rolland, Gustafson, Xiao, Whitehead, Berg, Lo, et~al.]{kirillov2023segment}
Alexander Kirillov, Eric Mintun, Nikhila Ravi, Hanzi Mao, Chloe Rolland, Laura Gustafson, Tete Xiao, Spencer Whitehead, Alexander~C Berg, Wan-Yen Lo, et~al.
\newblock Segment anything.
\newblock \emph{arXiv preprint arXiv:2304.02643}, 2023.

\bibitem[Kuhn(1955)]{kuhn1955hungarian}
Harold~W Kuhn.
\newblock The hungarian method for the assignment problem.
\newblock \emph{Naval research logistics quarterly}, 2\penalty0 (1-2):\penalty0 83--97, 1955.

\bibitem[Kweon et~al.(2023)Kweon, Yoon, and Yoon]{kweon2023weakly}
Hyeokjun Kweon, Sung-Hoon Yoon, and Kuk-Jin Yoon.
\newblock Weakly supervised semantic segmentation via adversarial learning of classifier and reconstructor.
\newblock In \emph{Proceedings of the IEEE/CVF Conference on Computer Vision and Pattern Recognition}, pages 11329--11339, 2023.

\bibitem[Lee et~al.(2021)Lee, Hyun, Seong, and Kim]{lee2021unsupervised}
Suhyeon Lee, Junhyuk Hyun, Hongje Seong, and Euntai Kim.
\newblock Unsupervised domain adaptation for semantic segmentation by content transfer.
\newblock In \emph{Proceedings of the AAAI conference on Artificial Intelligence}, pages 8306--8315, 2021.

\bibitem[Li et~al.(2023{\natexlab{a}})Li, Zhang, Sun, Zou, Liu, Yang, Li, Zhang, and Gao]{li2023semantic}
Feng Li, Hao Zhang, Peize Sun, Xueyan Zou, Shilong Liu, Jianwei Yang, Chunyuan Li, Lei Zhang, and Jianfeng Gao.
\newblock Semantic-sam: Segment and recognize anything at any granularity.
\newblock \emph{arXiv preprint arXiv:2307.04767}, 2023{\natexlab{a}}.

\bibitem[Li et~al.(2023{\natexlab{b}})Li, Wang, Cheng, Yu, Zhao, Song, Liu, Yuan, and Chen]{li2023acseg}
Kehan Li, Zhennan Wang, Zesen Cheng, Runyi Yu, Yian Zhao, Guoli Song, Chang Liu, Li Yuan, and Jie Chen.
\newblock Acseg: Adaptive conceptualization for unsupervised semantic segmentation.
\newblock In \emph{Proceedings of the IEEE/CVF Conference on Computer Vision and Pattern Recognition}, pages 7162--7172, 2023{\natexlab{b}}.

\bibitem[Li et~al.(2023{\natexlab{c}})Li, Xu, Liu, Tong, Lyu, and Zhou]{li2023semanticremote}
Xin Li, Feng Xu, Fan Liu, Yao Tong, Xin Lyu, and Jun Zhou.
\newblock Semantic segmentation of remote sensing images by interactive representation refinement and geometric prior-guided inference.
\newblock \emph{IEEE Transactions on Geoscience and Remote Sensing}, 2023{\natexlab{c}}.

\bibitem[Liao et~al.(2021)Liao, Kar, and Fidler]{liao2021towards}
Yuan-Hong Liao, Amlan Kar, and Sanja Fidler.
\newblock Towards good practices for efficiently annotating large-scale image classification datasets.
\newblock In \emph{Proceedings of the IEEE/CVF Conference on Computer Vision and Pattern Recognition}, pages 4350--4359, 2021.

\bibitem[Lin et~al.(2014)Lin, Maire, Belongie, Hays, Perona, Ramanan, Doll{\'a}r, and Zitnick]{lin2014microsoft}
Tsung-Yi Lin, Michael Maire, Serge Belongie, James Hays, Pietro Perona, Deva Ramanan, Piotr Doll{\'a}r, and C~Lawrence Zitnick.
\newblock Microsoft coco: Common objects in context.
\newblock In \emph{Computer Vision--ECCV 2014: 13th European Conference, Zurich, Switzerland, September 6-12, 2014, Proceedings, Part V 13}, pages 740--755. Springer, 2014.

\bibitem[Ling et~al.(2019)Ling, Gao, Kar, Chen, and Fidler]{ling2019fast}
Huan Ling, Jun Gao, Amlan Kar, Wenzheng Chen, and Sanja Fidler.
\newblock Fast interactive object annotation with curve-gcn.
\newblock In \emph{Proceedings of the IEEE/CVF conference on computer vision and pattern recognition}, pages 5257--5266, 2019.

\bibitem[Liu et~al.(2023)Liu, Zeng, Ren, Li, Zhang, Yang, Li, Yang, Su, Zhu, et~al.]{liu2023grounding}
Shilong Liu, Zhaoyang Zeng, Tianhe Ren, Feng Li, Hao Zhang, Jie Yang, Chunyuan Li, Jianwei Yang, Hang Su, Jun Zhu, et~al.
\newblock Grounding dino: Marrying dino with grounded pre-training for open-set object detection.
\newblock \emph{arXiv preprint arXiv:2303.05499}, 2023.

\bibitem[Medeiros(2023)]{LangSAM}
Luca Medeiros.
\newblock lang-segment-anything.
\newblock \url{https://github.com/luca-medeiros/lang-segment-anything}, 2023.

\bibitem[Melas-Kyriazi et~al.(2022)Melas-Kyriazi, Rupprecht, Laina, and Vedaldi]{melas2022deep}
Luke Melas-Kyriazi, Christian Rupprecht, Iro Laina, and Andrea Vedaldi.
\newblock Deep spectral methods: A surprisingly strong baseline for unsupervised semantic segmentation and localization.
\newblock In \emph{Proceedings of the IEEE/CVF Conference on Computer Vision and Pattern Recognition}, pages 8364--8375, 2022.

\bibitem[Radford et~al.(2021)Radford, Kim, Hallacy, Ramesh, Goh, Agarwal, Sastry, Askell, Mishkin, Clark, et~al.]{radford2021learning}
Alec Radford, Jong~Wook Kim, Chris Hallacy, Aditya Ramesh, Gabriel Goh, Sandhini Agarwal, Girish Sastry, Amanda Askell, Pamela Mishkin, Jack Clark, et~al.
\newblock Learning transferable visual models from natural language supervision.
\newblock In \emph{International conference on machine learning}, pages 8748--8763. PMLR, 2021.

\bibitem[Research(2023)]{GroundedSAM}
IDEA Research.
\newblock Grounded segment anything.
\newblock \url{https://github.com/IDEA-Research/Grounded-Segment-Anything}, 2023.

\bibitem[Rombach et~al.(2022)Rombach, Blattmann, Lorenz, Esser, and Ommer]{rombach2022high}
Robin Rombach, Andreas Blattmann, Dominik Lorenz, Patrick Esser, and Bj{\"o}rn Ommer.
\newblock High-resolution image synthesis with latent diffusion models.
\newblock In \emph{Proceedings of the IEEE/CVF conference on computer vision and pattern recognition}, pages 10684--10695, 2022.

\bibitem[Ru et~al.(2023)Ru, Zheng, Zhan, and Du]{ru2023token}
Lixiang Ru, Heliang Zheng, Yibing Zhan, and Bo Du.
\newblock Token contrast for weakly-supervised semantic segmentation.
\newblock In \emph{Proceedings of the IEEE/CVF Conference on Computer Vision and Pattern Recognition}, pages 3093--3102, 2023.

\bibitem[Scheibenreif et~al.(2022)Scheibenreif, Hanna, Mommert, and Borth]{scheibenreif2022self}
Linus Scheibenreif, Jo{\"e}lle Hanna, Michael Mommert, and Damian Borth.
\newblock Self-supervised vision transformers for land-cover segmentation and classification.
\newblock In \emph{Proceedings of the IEEE/CVF Conference on Computer Vision and Pattern Recognition}, pages 1422--1431, 2022.

\bibitem[Seong et~al.(2023)Seong, Moon, Lee, and Heo]{seong2023leveraging}
Hyun~Seok Seong, WonJun Moon, SuBeen Lee, and Jae-Pil Heo.
\newblock Leveraging hidden positives for unsupervised semantic segmentation.
\newblock In \emph{Proceedings of the IEEE/CVF Conference on Computer Vision and Pattern Recognition}, pages 19540--19549, 2023.

\bibitem[Strudel et~al.(2021)Strudel, Garcia, Laptev, and Schmid]{strudel2021segmenter}
Robin Strudel, Ricardo Garcia, Ivan Laptev, and Cordelia Schmid.
\newblock Segmenter: Transformer for semantic segmentation.
\newblock In \emph{Proceedings of the IEEE/CVF international conference on computer vision}, pages 7262--7272, 2021.

\bibitem[Tian et~al.(2024)Tian, Aggarwal, Colaco, Kira, and Gonzalez-Franco]{tian2024diffuse}
Junjiao Tian, Lavisha Aggarwal, Andrea Colaco, Zsolt Kira, and Mar Gonzalez-Franco.
\newblock Diffuse attend and segment: Unsupervised zero-shot segmentation using stable diffusion.
\newblock In \emph{Proceedings of the IEEE/CVF Conference on Computer Vision and Pattern Recognition}, pages 3554--3563, 2024.

\bibitem[Van~Gansbeke et~al.(2021)Van~Gansbeke, Vandenhende, Georgoulis, and Van~Gool]{van2021unsupervised}
Wouter Van~Gansbeke, Simon Vandenhende, Stamatios Georgoulis, and Luc Van~Gool.
\newblock Unsupervised semantic segmentation by contrasting object mask proposals.
\newblock In \emph{Proceedings of the IEEE/CVF International Conference on Computer Vision}, pages 10052--10062, 2021.

\bibitem[Vaswani et~al.(2017)Vaswani, Shazeer, Parmar, Uszkoreit, Jones, Gomez, Kaiser, and Polosukhin]{vaswani2017attention}
Ashish Vaswani, Noam Shazeer, Niki Parmar, Jakob Uszkoreit, Llion Jones, Aidan~N Gomez, {\L}ukasz Kaiser, and Illia Polosukhin.
\newblock Attention is all you need.
\newblock \emph{Advances in neural information processing systems}, 30, 2017.

\bibitem[Vobecky et~al.(2022)Vobecky, Hurych, Sim{\'e}oni, Gidaris, Bursuc, P{\'e}rez, and Sivic]{vobecky2022drive}
Antonin Vobecky, David Hurych, Oriane Sim{\'e}oni, Spyros Gidaris, Andrei Bursuc, Patrick P{\'e}rez, and Josef Sivic.
\newblock Drive\&segment: Unsupervised semantic segmentation of urban scenes via cross-modal distillation.
\newblock In \emph{European Conference on Computer Vision}, pages 478--495. Springer, 2022.

\bibitem[Wang et~al.(2023)Wang, Shen, Yuan, Du, Li, Hu, Crowley, and Vaufreydaz]{wang2023tokencut}
Yangtao Wang, Xi Shen, Yuan Yuan, Yuming Du, Maomao Li, Shell~Xu Hu, James~L Crowley, and Dominique Vaufreydaz.
\newblock Tokencut: Segmenting objects in images and videos with self-supervised transformer and normalized cut.
\newblock \emph{IEEE transactions on pattern analysis and machine intelligence}, 45\penalty0 (12):\penalty0 15790--15801, 2023.

\bibitem[Yan et~al.(2023)Yan, Liu, Liang, Wang, Li, and Wang]{yan2023semantic}
Jining Yan, Jingwei Liu, Dong Liang, Yi Wang, Jun Li, and Lizhe Wang.
\newblock Semantic segmentation of land cover in urban areas by fusing multi-source satellite image time series.
\newblock \emph{IEEE Transactions on Geoscience and Remote Sensing}, 2023.

\bibitem[Yan et~al.(2020)Yan, Li, Liu, Liu, and Chen]{yan2020roboseg}
Qingqing Yan, Shu Li, Chengju Liu, Ming Liu, and Qijun Chen.
\newblock Roboseg: Real-time semantic segmentation on computationally constrained robots.
\newblock \emph{IEEE Transactions on Systems, Man, and Cybernetics: Systems}, 52\penalty0 (3):\penalty0 1567--1577, 2020.

\bibitem[Yang et~al.(2023)Yang, Ma, Wen, Jiang, Yuan, and Zhu]{yang2023recognize}
Haosen Yang, Chuofan Ma, Bin Wen, Yi Jiang, Zehuan Yuan, and Xiatian Zhu.
\newblock Recognize any regions.
\newblock \emph{arXiv preprint arXiv:2311.01373}, 2023.

\bibitem[Yang and Gong(2024)]{yang2024foundation}
Xiaobo Yang and Xiaojin Gong.
\newblock Foundation model assisted weakly supervised semantic segmentation.
\newblock In \emph{Proceedings of the IEEE/CVF Winter Conference on Applications of Computer Vision}, pages 523--532, 2024.

\bibitem[Yao et~al.(2023)Yao, Guan, Huang, Li, Sha, Yue, Lim, Seo, Man, Zhu, et~al.]{yao2023radar}
Shanliang Yao, Runwei Guan, Xiaoyu Huang, Zhuoxiao Li, Xiangyu Sha, Yong Yue, Eng~Gee Lim, Hyungjoon Seo, Ka~Lok Man, Xiaohui Zhu, et~al.
\newblock Radar-camera fusion for object detection and semantic segmentation in autonomous driving: A comprehensive review.
\newblock \emph{IEEE Transactions on Intelligent Vehicles}, 2023.

\bibitem[Yin et~al.(2022)Yin, Wang, Wang, Xu, Zhang, Li, and Jin]{yin2022transfgu}
Zhaoyuan Yin, Pichao Wang, Fan Wang, Xianzhe Xu, Hanling Zhang, Hao Li, and Rong Jin.
\newblock Transfgu: a top-down approach to fine-grained unsupervised semantic segmentation.
\newblock In \emph{European conference on computer vision}, pages 73--89. Springer, 2022.

\bibitem[You et~al.(2023)You, Zhang, Gan, Du, Zhang, Wang, Cao, Chang, and Yang]{you2023ferret}
Haoxuan You, Haotian Zhang, Zhe Gan, Xianzhi Du, Bowen Zhang, Zirui Wang, Liangliang Cao, Shih-Fu Chang, and Yinfei Yang.
\newblock Ferret: Refer and ground anything anywhere at any granularity.
\newblock \emph{arXiv preprint arXiv:2310.07704}, 2023.

\bibitem[Yu et~al.(2022)Yu, Wang, Kim, Qiao, Collins, Zhu, Adam, Yuille, and Chen]{yu2022cmt}
Qihang Yu, Huiyu Wang, Dahun Kim, Siyuan Qiao, Maxwell Collins, Yukun Zhu, Hartwig Adam, Alan Yuille, and Liang-Chieh Chen.
\newblock Cmt-deeplab: Clustering mask transformers for panoptic segmentation.
\newblock In \emph{Proceedings of the IEEE/CVF conference on computer vision and pattern recognition}, pages 2560--2570, 2022.

\bibitem[Yu et~al.(2023)Yu, Zhao, Zhang, Zhang, Chen, Yang, Peng, and Zhang]{yu2023cross}
Ziqi Yu, Botao Zhao, Yipin Zhang, Shengjie Zhang, Xiang Chen, Haibo Yang, Tingying Peng, and Xiao-Yong Zhang.
\newblock Cross-grained contrastive representation for unsupervised lesion segmentation in medical images.
\newblock In \emph{Proceedings of the IEEE/CVF International Conference on Computer Vision}, pages 2347--2354, 2023.

\bibitem[Zadaianchuk et~al.(2022)Zadaianchuk, Kleindessner, Zhu, Locatello, and Brox]{zadaianchuk2022unsupervised}
Andrii Zadaianchuk, Matthaeus Kleindessner, Yi Zhu, Francesco Locatello, and Thomas Brox.
\newblock Unsupervised semantic segmentation with self-supervised object-centric representations.
\newblock \emph{arXiv preprint arXiv:2207.05027}, 2022.

\bibitem[Zhang et~al.(2023{\natexlab{a}})Zhang, Li, Li, Huang, Huang, and Zhang]{zhang2023rethinking}
Daoan Zhang, Chenming Li, Haoquan Li, Wenjian Huang, Lingyun Huang, and Jianguo Zhang.
\newblock Rethinking alignment and uniformity in unsupervised image semantic segmentation.
\newblock In \emph{Proceedings of the AAAI Conference on Artificial Intelligence}, pages 11192--11200, 2023{\natexlab{a}}.

\bibitem[Zhang et~al.(2023{\natexlab{b}})Zhang, Zhou, Mai, Hu, Guan, Li, and Mu]{zhang2023text2seg}
Jielu Zhang, Zhongliang Zhou, Gengchen Mai, Mengxuan Hu, Zihan Guan, Sheng Li, and Lan Mu.
\newblock Text2seg: Remote sensing image semantic segmentation via text-guided visual foundation models.
\newblock \emph{arXiv preprint arXiv:2304.10597}, 2023{\natexlab{b}}.

\bibitem[Zhao et~al.(2024)Zhao, Mithun, Rajvanshi, Chiu, and Samarasekera]{zhao2024unsupervised}
Xingchen Zhao, Niluthpol~Chowdhury Mithun, Abhinav Rajvanshi, Han-Pang Chiu, and Supun Samarasekera.
\newblock Unsupervised domain adaptation for semantic segmentation with pseudo label self-refinement.
\newblock In \emph{Proceedings of the IEEE/CVF Winter Conference on Applications of Computer Vision}, pages 2399--2409, 2024.

\bibitem[Ziegler and Asano(2022)]{ziegler2022self}
Adrian Ziegler and Yuki~M Asano.
\newblock Self-supervised learning of object parts for semantic segmentation.
\newblock In \emph{Proceedings of the IEEE/CVF Conference on Computer Vision and Pattern Recognition}, pages 14502--14511, 2022.

\end{thebibliography}
}


\end{document}